\documentclass[lettersize,journal]{IEEEtran}
\usepackage{amsmath,amsfonts}
\usepackage{algorithm}
\usepackage{array}
\usepackage[caption=false,font=normalsize,labelfont=sf,textfont=sf]{subfig}
\usepackage{textcomp}
\usepackage{stfloats}
\usepackage{url}
\usepackage{verbatim}
\usepackage{graphicx}
\usepackage{cite}

\usepackage{amsmath}
\usepackage{amssymb}  
\usepackage{enumitem}
\usepackage{algpseudocode}
\usepackage{array}
\usepackage{tabularx}
\usepackage{booktabs}
\usepackage{multirow} 
\usepackage{graphicx}
\usepackage{adjustbox}
\usepackage[table]{xcolor} 
\usepackage{url} 
\definecolor{lightgray}{RGB}{240,240,240} 

\usepackage{pifont}
\newcommand{\cmark}{\ding{51}} 
\newcommand{\xmark}{\ding{55}}
\begin{document}

\title{Enhancing Vision-Language Navigation with Multimodal Event Knowledge from Real-World Indoor Tour Videos}
\author{Haoxuan~Xu,
        Tianfu~Li,
        Wenbo~Chen,
        Yi~Liu,
        Xingxing Zuo,
        Yaoxian~Song*
        and~Haoang~Li*\thanks{*Corresponding author.}%
\thanks{Haoxuan Xu, Tianfu Li, Wenbo Chen and Haoang Li are with The Hong Kong University of Science and Technology (Guangzhou), Guangzhou 511453, China (e-mail: hxu095@connect.hkust-gz.edu.cn; tli794@connect.hkust-gz.edu.cn; wchen361@connect.hkust-gz.edu.cn;hlei573@connect.hkust-gz.edu.cn; haoang.li.cuhk@gmail.com).}%
\thanks{Yi Liu is with Tsinghua University, Shenzhen 518055, China (e-mail: yiliu24@mails.tsinghua.edu.cn).}%
\thanks{Xingxing Zuo is with Mohamed Bin Zayed University of Artificial Intelligence (MBZUAI), Abu Dhabi 31122, UAE, (e-mail: xingxing.zuo@mbzuai.ac.ae)}%
\thanks{Yaoxian Song is with Hangzhou City University, Hangzhou 310015, China (e-mail: songyaoxian@westlake.edu.cn).}%
}

\maketitle

\begin{abstract}
Vision-Language Navigation (VLN) agents often struggle with long-horizon reasoning in unseen environments, particularly when facing ambiguous, coarse-grained instructions. 
While recent advances use knowledge graph to enhance reasoning, the potential of multimodal event knowledge inspired by human episodic memory remains underexplored.
In this work, we propose an event-centric knowledge enhancement strategy for automated process knowledge mining and feature fusion to solve coarse-grained instruction and long-horizon reasoning in VLN task. 
First, we construct \textbf{YE-KG}, the first large-scale multimodal spatiotemporal knowledge graph, with over $86$k nodes and $83$k edges, derived from real-world indoor videos. By leveraging multimodal large language models (i.e., LLaVa, GPT4), we extract unstructured video streams into structured semantic-action-effect events to serve as explicit episodic memory.
Second, we introduce \textbf{STE-VLN}, which integrates the above graph into VLN models via a Coarse-to-Fine Hierarchical Retrieval mechanism. This allows agents to retrieve causal event sequences and dynamically fuse them with egocentric visual observations.
Experiments on REVERIE, R2R, and R2R-CE benchmarks demonstrate the efficiency of our event-centric strategy, outperforming state-of-the-art approaches across diverse action spaces.
Our data and code are available on the project website \url{https://sites.google.com/view/y-event-kg/}.


\end{abstract}


\section{Introduction}
\IEEEPARstart{V}ision‑Language Navigation (VLN)~\cite{anderson2018vision,ku2020room,qi2020reverie} seeks to empower embodied agents to autonomously traverse unseen indoor environments and reach designated targets guided by natural language instructions. Embedding such language‑guided navigation capabilities into robotic systems offers an intuitive mode of human–robot interaction for sectors like in‑home assistance~\cite{he2023mlanet}, warehouse logistics~\cite{liang2015automated}, and healthcare service~\cite{silvera2024robotics}, thereby significantly alleviating the operational burden on end users.

\begin{figure}[t!]
    \centering
    \includegraphics[width=0.98\linewidth]{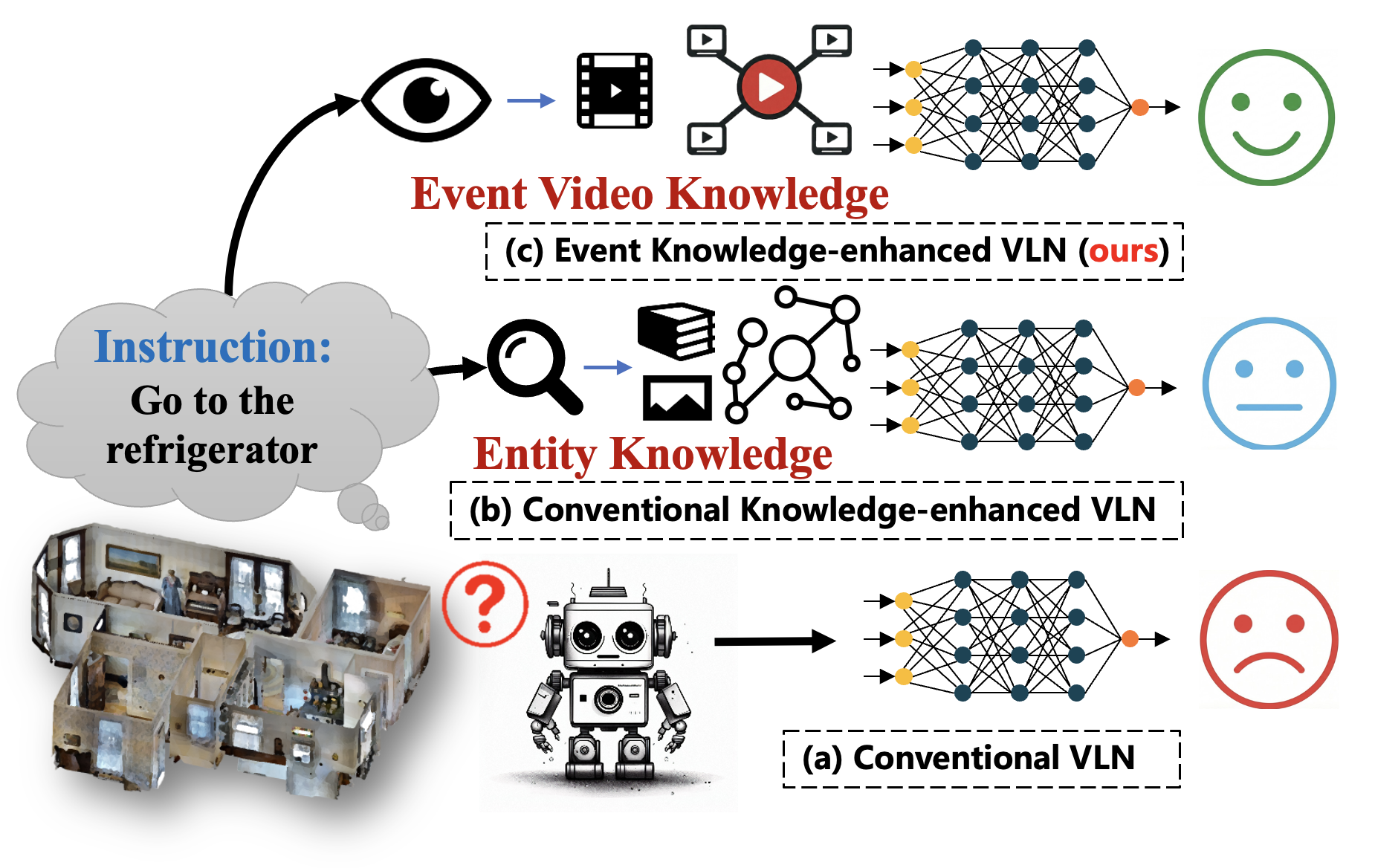}
    \caption{
      \textbf{Various VLN modeling paradigms.}
      \textbf{(a) Conventional VLN methods} design a planner based directly on instructions and visual observations~\cite{hong2021vln}, often lacking a deep understanding of complex indoor environmental knowledge.
      \textbf{(b) Previous knowledge-enhanced VLN models} introduce entity knowledge embeddings~\cite{song2024scene} as priors but fail to effectively associate objects and scenes with dynamic navigation actions.
      \textbf{(c) Our method} constructs video-based event knowledge for the VLN planner. It empowers the robot with explicit knowledge priors that uniquely capture the spatiotemporal features linking objects, scenes, and navigation actions.
    }
    \label{fig:VLN_modeling_paradigms}
\end{figure}

The vision-language navigation (VLN) task is typically formulated as a supervised multimodal sequence-to-sequence decision problem~\cite{hong2021vln,pashevich2021episodic}. Given natural language instructions and a temporally ordered sequence of egocentric visual observations, the agent aims to generate a sequence of actions to find the target area. Key challenges include abstract instruction understanding, multimodal perception, and sequential decision making. 
VLN architectures have evolved from early RNNs~\cite{dang2022unbiased,he2023mlanet} and Transformers~\cite{chen2021history,chen2022think,du2024delan} to recent LLM-based zero-shot planners~\cite{qiao2024llm,zhou2024navgpt,chen2025llm,wang2025open}. However, as illustrated in Figure~\ref{fig:VLN_modeling_paradigms}(a), despite architectural evolutions, a critical cognitive gap persists. Mainly parametric models follow a reactive paradigm, relying on visual pattern matching while ignoring external priors of indoor layouts and object-room relationships. This deficiency hampers generalization in unseen environments, particularly under coarse-grained instructions. In contrast, human navigation is a predictive process driven by episodic memory~\cite{tulving1972episodic}, which leverages past experiences to anticipate unobserved areas rather than merely reacting to visual inputs. To mimic this capability, incorporating explicit knowledge graphs as structural priors is essential.

Incorporating knowledge graphs (Fig.~\ref{fig:VLN_modeling_paradigms}(b)) offers a promising pathway to mimic such cognitive capabilities, yet prior efforts suffer from critical limitations regarding content and modality.
First, most knowledge graphs like ConceptNet~\cite{liu2004conceptnet} and Scene-KG~\cite{song2024scene} are entity-centric and static. They fail to capture process knowledge linking actions to scenes, leaving agents unable to decompose coarse-grained instructions into executable steps. This deficiency directly leads to planning failures, causing agents to wander aimlessly.
Second, existing event-based attempts~\cite{zhao2024towards} are typically restricted to unimodal text or simulated environments. Lacking real-world visual cues, these methods struggle to align abstract textual plans with dynamic visual observations. This modality gap prevents agents from recognizing visual triggers in the physical world, resulting in execution errors during navigation.

To address the above limitations in content and modality, we propose a framework for automated process knowledge mining and spatiotemporal feature fusion. To equip the agent with explicit procedural priors, we construct the \textbf{Y}ouTube-\textbf{E}vent \textbf{K}nowledge \textbf{G}raph (YE-KG), the first multimodal event knowledge graph designed for VLN task. We achieve this process in three aspects. 
First, we curate a massive dataset of over 320 hours of real-world indoor tour videos, ensuring diverse coverage of layout dynamics. 
Second, leveraging the rich prior knowledge of large multimodal models, we employ LLaVA-Video~\cite{zhang2024video} to parse unstructured video streams into structured ``Semantic-Action-Effect'' events, followed by GPT-4 for refinement to mitigate hallucinations. This step extracts the causal logic of navigation (e.g., entering a kitchen implies approaching a fridge) while filtering out hallucinations. 
Third, we structure these events into a directed graph with over 86k nodes and 83k edges, serving as a learnable episodic memory that captures explicit room-object transitions.

For the effective integration of the above knowledge into VLN task, we propose the \textbf{S}patio-\textbf{T}emporal \textbf{E}vent-enhanced \textbf{V}ision-\textbf{L}anguage \textbf{N}avigation (STE-VLN) framework. To bridge the gap between abstract instructions and visual events, we design a Coarse-to-Fine Hierarchical Retrieval mechanism. 
For global planning, the coarse retrieval stage constructs a topological sub-graph from textual events to prevent aimless wandering. 
For local grounding, the fine retrieval stage provides visual foresight by recalling specific video clips. 
Moreover, to merge these multimodal inputs, we introduce an Adaptive Spatio-Temporal Feature Fusion (ASTFF) mechanism. We explicitly concatenate causally linked future video features into the visual stream and append event descriptions to instructions. This strategy aligns current observations with historical priors, shifting the paradigm from reactive matching to predictive reasoning. For inference efficiency, our method leverages the RAG~\cite{lewis2020rag} scheme. By optimizing the feature indexing, we enable the agent to access relevant priors with an ultra-low latency of 0.02 ms. This ensures that the incorporation of external knowledge imposes negligible computational burden, making it highly suitable for real-time robotic deployment.

In summary, the contributions of this paper are as follows:

\begin{enumerate}
    \item We construct YE-KG, the first large-scale multimodal knowledge graph mined from over 320 hours of open-world videos. Comprising more than 86k nodes and 83k edges, it provides rich spatiotemporal process priors to bridge the gap between static entities and dynamic navigation.
    
    \item We propose STE-VLN, a knowledge-enhanced framework built on YE-KG. It features a coarse-to-fine retrieval mechanism and an adaptive fusion module to dynamically align global textual plans with local visual foresight.
    
    \item Extensive experiments on three widely used benchmarks, REVERIE~\cite{qi2020reverie}, R2R~\cite{anderson2018vision}, and R2R-CE~\cite{krantz2020beyond}, together with a real-world robotic validation, demonstrate that our method consistently outperforms state-of-the-art baselines and exhibits robust sim-to-real generalization.
\end{enumerate}

\section{Related work}

\subsection{Vision-Language Navigation}
VLN requires an embodied agent to interpret natural language instructions and traverse unseen environments. This task is generally categorized into navigation on discrete graphs~\cite{anderson2018vision} and continuous 3-D spaces~\cite{krantz2020beyond}.
Early approaches treated VLN as a sequence-to-sequence problem using RNNs~\cite{anderson2018vision, an2021neighbor} or auxiliary reasoning objectives~\cite{dou2022foam, wang2019reinforced}. The advent of Transformers shifted the focus to global context modeling. HAMT~\cite{chen2021history} and HOP+~\cite{qiao2023hop+} introduced history-aware mechanisms to capture long-range dependencies. To enhance spatial reasoning, DUET~\cite{chen2022think} and DSRG~\cite{wang2023dual} construct topological maps for global planning, while TD-STP~\cite{zhao2022target} focuses on target-driven structured modeling. Similarly, BEVBert~\cite{an2022bevbert}, GridMM~\cite{wang2023gridmm}, and GeoVLN~\cite{huo2023geovln} project egocentric observations into bird’s-eye view (BEV) or geometric grids to improve cross-modal alignment. Furthermore, methods like GELA~\cite{Cui_2023_ICCV} and EnvEdit~\cite{li2022envedit} utilize entity grounding and environment editing to boost robustness. Recently, GOAT~\cite{wang2024vision} integrated object-aware tokens to improve fine-grained recognition.
Navigating in continuous environments (R2R-CE) presents unique challenges in low-level control. Early methods like Seq2Seq~\cite{krantz2020beyond}, LAW~\cite{raychaudhuri2021language}, and CMTP~\cite{chen2021topological} applied sequence modeling and topological mapping to continuous spaces. HPN~\cite{krantz2021waypoint} and SASRA~\cite{irshad2021sasra} focused on waypoint prediction and semantic reasoning. To bridge the domain gap, Sim2Sim~\cite{krantz2022sim} and CWP~\cite{hong2022bridging} transferred graph-based policies to continuous settings. Most recently, ETPNav~\cite{an2024etpnav} evolved topological plans online to handle long-horizon traversal.
With the rise of Large Language Models (LLMs), works like NavGPT~\cite{zhou2024navgpt}, MapGPT~\cite{chen2024mapgpt}, and NaviLLM~\cite{zheng2024towards} utilize zero-shot reasoning for high-level planning. 
However, these agents often lack explicit grounding in physical dynamics, leading to hallucinations. This limitation underscores the necessity of incorporating external structured knowledge to guide robust decision-making.

\subsection{Knowledge Enhancement in Vision-Language Navigation}

Since VLN datasets are notoriously expensive to collect, incorporating external knowledge has become a critical strategy to mitigate data scarcity and equip agents with common sense. Speaker models~\cite{dou2022foam} generate synthetic instructions for random trajectories, while EnvEdit~\cite{li2022envedit} alters environmental textures and styles to diversify training scenarios. Recent works like FLAME~\cite{xu2025flame} utilize synthetic data to enhance the adaptation capabilities of Multimodal LLMs. While effective for visual diversity, these methods rarely introduce new behavioral or procedural knowledge that is absent in the original simulator. Structured priors from knowledge graphs provide rich semantic context. Early methods leveraged generic commonsense bases~\cite{liu2025kg} like ConceptNet~\cite{liu2004conceptnet} to associate objects with rooms~\cite{gao2021room, wang2025goal, gao2025visual}. More recently, Scene-driven Multimodal KGs~\cite{song2024scene} were proposed to align fine-grained visual features with textual entities. However, these approaches predominantly model static entity relationships (Object-Room). They fail to capture the spatiotemporal and causal nature of navigation events (Action-Effect). The most related work, VLN-EventKG~\cite{zhao2024towards}, attempts to extract events but relies solely on textual data from existing simulators. 
This highlights a critical gap: existing KGs lack the dynamic, spatiotemporal richness found in the real world. To address this, we turn to massive open-world video data resources as a source for mining event-level process knowledge.

\subsection{Learning from Open-World Data}

Recent trends in Artificial Intelligence have shifted from training on small, curated datasets to mining generalizable knowledge from massive, open-world sources. 
In Computer Vision and NLP, exploiting web-scale image-text pairs~\cite{radford2021learning, jia2021scaling} and self-supervised visual patterns~\cite{he2022masked, caron2021emerging} has unlocked zero-shot capabilities. 
Beyond static images, researchers leverage large-scale video datasets~\cite{kay2017kinetics, goyal2017something} and egocentric collections~\cite{grauman2022ego4d, damen2018scaling} to train robust representations for robotic perception, exemplified by VIP~\cite{ma2022vip} and VC-1~\cite{majumdar2023we}.
In Embodied AI, this paradigm facilitates generalist policy learning~\cite{o2024open}, where VLA models like RT-2~\cite{zitkovich2023rt} and navigation foundation models like NaVila~\cite{cheng2024navila} transfer web knowledge to control and traversability estimation~\cite{chang2020semantic, lin2023learning}.
However, these methods typically encode knowledge \textit{implicitly} within ``black-box'' neural weights, hindering interpretability and precise reasoning. 
In contrast, we mine this video data into an explicit spatiotemporal knowledge graph (\textbf{YE-KG}), combining web-scale generalizability with the structured symbolic reasoning advocated in knowledge-enhanced approaches.

\section{Multimodal Event Knowledge Graph Construction}  

\begin{figure*}[t]
    \centering
    \includegraphics[width=0.95\linewidth]{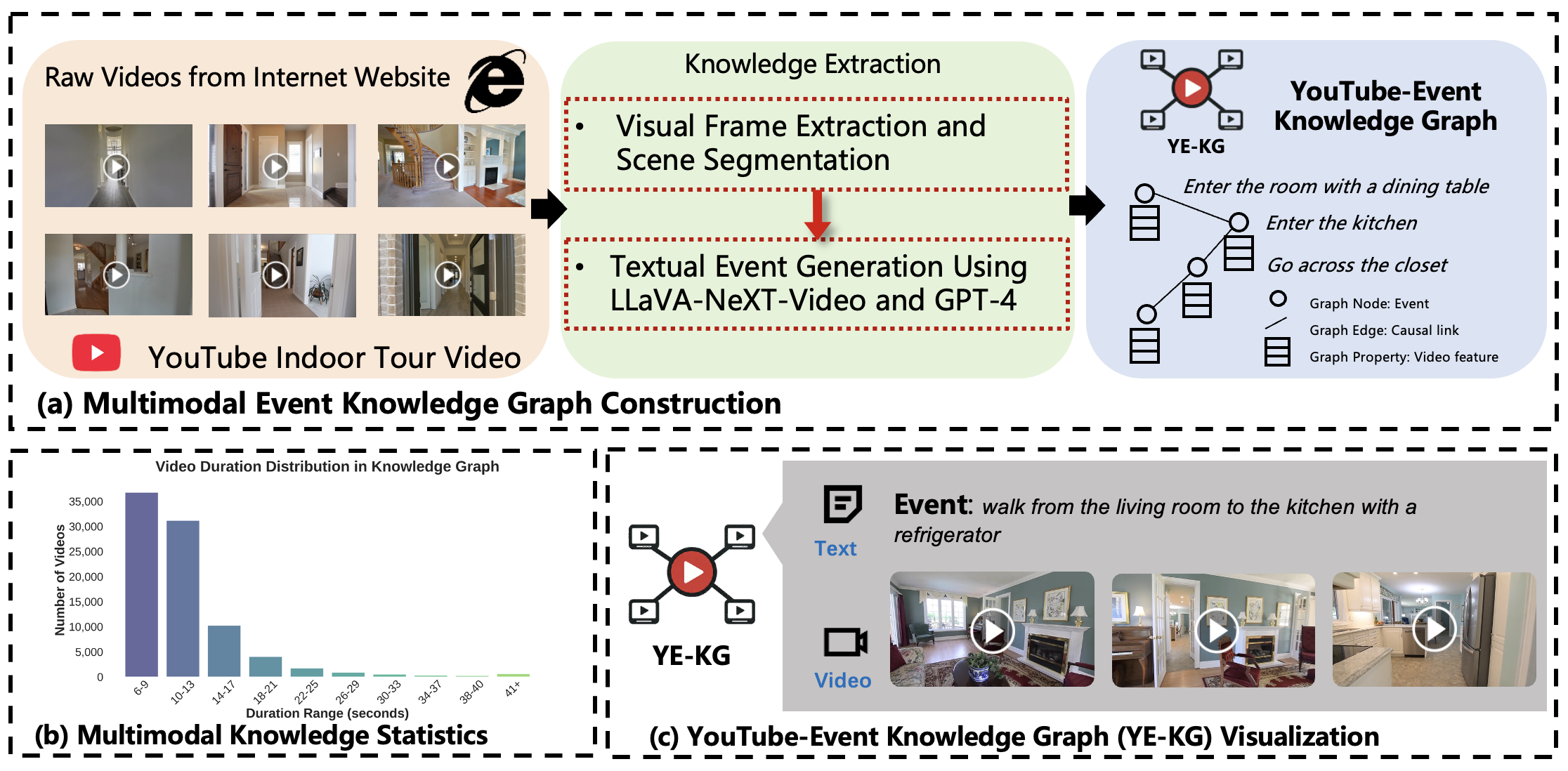}
    \caption{\textbf{Overview of YouTube‑Event‑KG (YE-KG)}
    \textbf{(a) Multimodal Event Knowledge Graph Construction.}
    We first collect large-scale real-world indoor tour videos from YouTube, followed by visual frame extraction and scene segmentation. Then, we employ LLaVA-NeXT-Video together with GPT-4 to generate semantically grounded event descriptions as nodes, connecting them via directed edges representing temporal adjacency (causal navigational links) to yield the \textbf{YE-KG}.  
    \textbf{(b) Data Statistics.}
    The video duration distribution in the knowledge graph is tightly concentrated between 6 and 13 seconds, highlighting the corpus’s high quality and temporal consistency.
    \textbf{(c) YE‑KG Visualization.}
    An example event is visualized together with its associated video features, illustrating how event nodes are grounded in spatiotemporal visual context.}
    \label{fig:Event_Multimodal_Knowledge_Graph_Construction}
\end{figure*}

We propose the first multimodal event knowledge construction method and build the first Event-level VLN knowledge Graph from indoor tour videos of rental properties on YouTube (\textbf{YE-KG}) for vision-language navigation as shown in Fig.~\ref{fig:Event_Multimodal_Knowledge_Graph_Construction}, which explicitly encodes spatiotemporal features and sequence event relations from real-world exploration sequences. 
To ensure data quality and model robustness, YE-KG is built with a carefully controlled entity distribution. Empirical analyses confirm that the graph lacks a pronounced long-tail pattern, shown in Fig.~\ref{fig:Event_Multimodal_Knowledge_Graph_Construction}(b), enabling more stable learning across nodes.
By encoding both action events and contextual scenes, YE-KG provides a rich, hierarchically grounded knowledge base for agents to reason about long-horizon navigation tasks in complex indoor environments.

\subsection{Formulation of Spatiotemporal Events}
To formalize the navigation process, we define the navigation event as a multimodal spatiotemporal transition unit. Formally, an event $e$ is defined as a tuple:
\begin{equation}
e = (\mathcal{R}_{\text{src}}, \mathcal{A}, \mathcal{R}_{\text{tgt}}, C_{\text{scene}}, V_{\text{clip}}, T_{\text{desc}}),
\end{equation}
where $\mathcal{R}_{\text{src}}$ and $\mathcal{R}_{\text{tgt}}$ denote the semantic regions (e.g., bedroom) of the source and target, $\mathcal{A}$ represents the high-level action (e.g., walk across the corridor) and $C_{\text{scene}}$ encodes the static context. 
Furthermore, $V_{\text{clip}}$ and $T_{\text{desc}}$ are the visual representations and textual descriptions mined from real-world videos (detailed in Sec. \ref{sec:3_B} and Sec. \ref{sec:3_C}, respectively).

\subsection{Event Extraction from Open-World Videos}
\label{sec:3_B}
To instantiate the event definition above with real-world data, we establish a pipeline to mine visual event clips from open-world sources. Inspired by prior work leveraging real-world videos for navigation learning ~\cite{chang2020semantic, lin2023learning}, we collect 3,471 high-quality real estate tour videos (over 320 hours total) from YouTube. These in-the-wild videos capture diverse indoor layouts and natural navigation behaviors, forming a rich basis for modeling navigational events beyond static scene representations.


Once the raw video is obtained, we perform a coarse-to-fine extraction to isolate meaningful indoor transitions. First, each video is sampled at 0.5 FPS to generate a frame sequence \(V_\text{YouTube}^i\). We apply a cleaning process \(E_{\text{frame}}\) to discard irrelevant frames (e.g., outdoor scenes), resulting in the indoor frame set \( \mathcal{V}_{\text{indoor}} \).

Given that navigation events are characterized by topological changes between functional regions (e.g., entering a kitchen from a living room), we require a segmentation method capable of understanding open-world semantic concepts. To this end, we employ CLIP~\cite{radford2021learning} to label frames based on their semantic content.
Formally, given a set of text prompts \(\mathcal{L}=\{l_k\}_{k=1}^K\) for room categories and a frame \( v \), we compute the probability distribution \(P\) via softmax-normalized cosine similarity: \( P(k|v) \propto \exp( \text{sim}(E_{\text{img}}(v), E_{\text{txt}}(l_k)) / \tau ) \). The label with the highest probability, \(\hat{l} = \arg\max_k P(k|v)\), is assigned to the frame, denoted as the \( E_{\text{clip\_room}} \) process.

After obtaining frame-level labels, temporally adjacent frames sharing the same label \(\hat{l}\) are merged into a single segment. To distill each segment into a canonical visual representation, we adopt the entropy-based selection strategy. Specifically, we identify the representative frame \( t_j \) that possesses the minimum Shannon entropy in its prediction probability \(P\), thereby selecting the view with the highest semantic certainty.
The sequence of frames between two successive representative frames, \( t_j \) and \( t_{j+1} \) (where \(\hat{l}_{t_j} \neq \hat{l}_{t_{j+1}}\)), constitutes an event clip \( V_{\text{event}}^j \). This process ensures that each extracted clip captures a distinct room-to-room transition. This temporal segmentation and clip generation process is formalized as \( E_{\text{clip\_event}} \). The entire pipeline is summarized as:
\begin{equation}\label{eq:frame_extraction}
\begin{aligned}
&{{\mathcal{V}}_{\textnormal{indoor}}} = {{E}_{\textnormal{frame}}}\left( {{\mathcal{V}_\textnormal{YouTube}}} \right),\\
&{{\mathcal{V}}_{\textnormal{room}}} = \left\{ {{{E} _{\textnormal{clip\_room}}}\left( {V_{\textnormal{indoor}}^i} \right)\mid V_{\textnormal{indoor}}^i \in {{\mathcal{V}}_{\textnormal{indoor}}}} \right\}, \\
&{{\mathcal{V}}_{\textnormal{event}}} = \left\{ {{{E} _{\textnormal{clip\_event}}}\left( {V_{\textnormal{room}}^i} \right)\mid V_{\textnormal{room}}^i \in {{\mathcal{V}}_{\textnormal{room}}}} \right\}.\\
\end{aligned} 
\end{equation}

This procedure is applied to the collected corpus, resulting in over 83k video clips (mean length \(\sim 11.6 \text{s}\)), which form the visual foundation for YE-KG.

\subsection{Semantically-Grounded Textual Generation}
\label{sec:3_C}
While visual clips provide spatiotemporal context, high-level planning requires explicit semantic understanding. Therefore, we generate textual descriptions using the vision–language model LLaVA-NeXT-Video~\cite{zhang2024video}. For each event clip \(V_{\text{event}}^{i} \in V_{\text{event}}\), the model produces an initial description \(T_{\text{event\_raw}}\) following Eq.~\eqref{eq:llava_video}.
\begin{equation}\label{eq:llava_video}
\small
\begin{aligned}
{{\mathcal{T}}_{\textnormal{event\_raw}}} = \left\{ {\text{LLaVa-video}\left( {V_{\textnormal{event}}^i} \right)\mid V_{\textnormal{event}}^i \in {{\mathcal{V}}_{\textnormal{event}}}} \right\},
\end{aligned} 
\end{equation}
where \(\text{LLaVa-Video}: V_{\text{event}}\!\rightarrow\!\text{Text}\) maps a video segment to a natural-language string. As generative models can hallucinate details, we pass each raw description to GPT-4 for refinement and two-tier verification:
\begin{equation}
\begin{aligned}
{{\mathcal{T}}_{\textnormal{event}}} = \left\{ {\text{GPT-4}\left( {T_{\textnormal{event}}^i} \right)\mid T_{\textnormal{event}}^i \in {{\mathcal{T}}_{\textnormal{event\_raw}}}} \right\},
\end{aligned}
\end{equation}
where \(\text{GPT-4}\) outputs the final description \(T_{\text{event}}^i\) and assigns a label \(y_i \in \{\texttt{Event-0}, \texttt{Scene-1}\}\). Concretely, for each final description (denoted \(\mathcal{D}_i = T_{\text{event}}^i\)), if \(\mathcal{D}_i\) contains a clear \textit{[Source Room] \(\rightarrow\) [Action] \(\rightarrow\) [Target Room]} transition, we label the node as \texttt{Event-0}; if it only mentions static scene details (e.g., “The room has red curtains and a wooden table …”), we label it as \texttt{Scene-1}. Descriptions labeled \texttt{Scene-1}, while not depicting movement, are retained as contextual scene nodes to enrich environment details, and the verification step filters out hallucinated transitions to ensure event nodes correspond to genuine navigational steps. Finally, we construct the corpus as a set of verified event–text pairs:
\begin{equation}
\begin{aligned}
&{\text{YouTube-Event-Knowledge}} = \\ 
\quad &\left\{ {\left( {V_{\textnormal{event}}^i,T_{\textnormal{event}}^i} \right),i = 1,\dots,\left| {{{\mathcal{V}}_{\textnormal{event}}}} \right|} \right\}.
\end{aligned}
\end{equation}

\subsection{Graph Construction and Formalization}
Finally, we construct the YE-KG as a directed graph \(\mathcal{G} = (\mathcal{V}, \mathcal{E}, \mathcal{F})\), where each event node \(v_i = T_{\text{event}}^{i} \in \mathcal{V}\) is in one-to-one correspondence with YouTube-Event-Knowledge, and its visual feature \(f_i \in \mathcal{F}\) is extracted from the associated clip \(V_{\text{event}}^{i}\) using a Vision Transformer (ViT)~\cite{dosovitskiy2020image}. Temporal succession is encoded as edges \(e_{ij} \in \mathcal{E}\): a directed edge is established from \(v_i\) to \(v_j\) if \(T_{\text{event}}^{i}\) and \(T_{\text{event}}^{j}\) originate from the same raw video and \(T_{\text{event}}^{i}\) temporally precedes \(T_{\text{event}}^{j}\) (i.e., they are adjacent in the segmentation order), indicating a plausible navigational progression. The resulting YE-KG comprises over $83$k spatiotemporal relations.

\begin{figure*}[!t]
    \centering
    \includegraphics[width=0.98\linewidth]{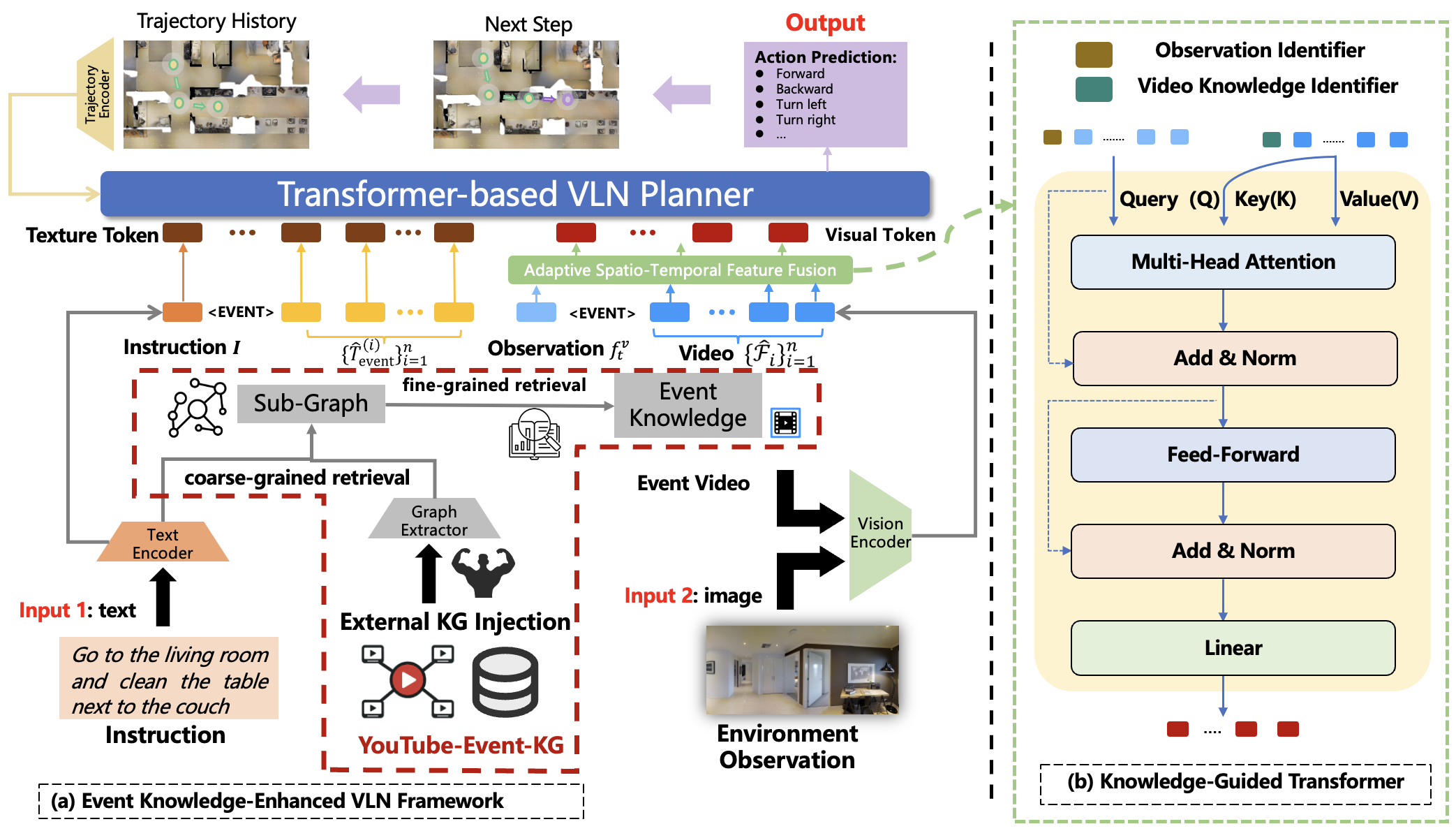}
    \caption{
    The overview of STE-VLN framework enhanced by the multimodal event knowledge. 
    (a) The texture instruction and egocentric visual observation are fed into the VLN planner, enriched by external knowledge. 
    The \textbf{red dashed line frame} illustrates the coarse-to-fine retrieval process: $\{\hat{T}_{\text{event}}^{(i)}\}_{i=1}^n$ and $\{\hat{\mathcal{F}}_i\}_{i=1}^{n}$ denote the top-$n$ retrieved textual event descriptions and their corresponding video features from the YE-KG, which are fused with the original instruction $I$ and current visual observation $f_t^v$, respectively.
    (b) Knowledge-Guided Transformer is designed to implement ASTFF, where the egocentric observation serves as the Query (Q) and retrieved video knowledge provides the Key (K) and Value (V).
    }
    \label{fig:pipeline}
\end{figure*}

\section{Spatiotemporal Event-Enhanced Navigation}
\label{sec:method}  

Building upon the \textbf{YE-KG}, which encodes rich spatiotemporal knowledge from real-world videos, our goal is to effectively integrate this explicit process knowledge into the navigation policy. 
To this end, we propose a novel multimodal feature fusion framework for vln task named \textbf{STE-VLN}. As illustrated in Fig.~\ref{fig:pipeline}, this framework is designed to enable robots to interpret coarse-grained instructions and reach targets in unfamiliar environments by bridging the gap between static instructions and dynamic execution.
Specifically, our approach introduces two key innovations: 
1) a Coarse-to-Fine Hierarchical Retrieval mechanism that hierarchically aligns the semantics of instructions with relevant visual context and event sequences in the YE-KG; 
and 2) an ASTFF module that efficiently fuses the retrieved knowledge into the agent's egocentric navigation representations.

\subsection{Coarse-to-Fine Hierarchical Retrieval}

VLN agents often struggle with coarse-grained instructions, as such directives lack the step-by-step details necessary for precise action selection. To address this challenge, we introduce the coarse-to-fine hierarchical retrieval mechanism, which dynamically leverages our YE-KG to guide the agent more effectively towards the target area. This mechanism is composed of two sequential stages: 1. Coarse-grained sub-graph construction and 2. Fine-grained contextual fusion during navigation.

\paragraph{Stage 1: Coarse-grained Sub-graph Construction}
Given a coarse instruction \(I\), we retrieve a compact set of relevant event experiences from YE-KG. We encode \(I\) with a pre-trained sentence encoder and run FAISS~\cite{johnson2019billion} vector search over $\mathcal{T}_\text{event}$. The top-\(K\) nearest instructions are retained together with their event sequences (textual descriptions and visual features), preserving the causal order of navigation.


Formally, we score each event description $T_{\textnormal{event}}^{\,i}$ in $\mathcal{T}_\textnormal{event}$ against $I$. To obtain the retrieval seeds, we seek a subset $\mathcal{S}$ of size $K$ that maximizes the cumulative similarity score with the instruction. The resulting seed set $\mathcal{T}_{\textnormal{seed}}$, is defined as:

\begin{equation} 
\label{eq:coarse_topk} 
\begin{aligned} \mathcal{T}_{\textnormal{seed}} = \underset{\substack{\mathcal{S} \subset \mathcal{T}_{\textnormal{event}}, \ |\mathcal{S}|=K}}{\arg\max} \sum_{T \in \mathcal{S}} \operatorname{sim}(I, T).
\end{aligned} 
\end{equation}

We then expand $\mathcal{T}_{\textnormal{seed}}$ by tracing the causal links in the graph: all events reachable via directed causal edges from the seeds are included. These expanded events, together with their associated video features, constitute the final event subset $\mathcal{T}_{\text{event\_sub}}$. This subset forms the nodes of the coarse subgraph $\mathcal{G}_{\text{sub}}$, providing the spatiotemporal contextual prior for instruction $I$.



\paragraph{Stage 2: Fine-grained Knowledge Retrieval}

Given the coarse subgraph $\mathcal{G}_{\text{sub}}$, at each timestep $t$, the agent encodes the current panoramic observation into a visual feature $f_t^v$ and queries $\mathcal{G}_{\text{sub}}$ to retrieve the top-$n$ visually similar events, each associated with historical visual features and textual descriptions.
By following the causal edges from these events to their subsequent nodes, the agent extracts predictive features like future scenes, enabling the prediction of what typically follows the current view and thus facilitating long-horizon decision making.

To score similarity, we adopt an RAG~\cite{lewis2020rag} scheme. Let \(\mathcal{F}_{\text{sub}} = \{\mathbf{f}_j\}_{j=1}^{|\mathcal{V}_{\text{sub}}|}\) denote the set of video embeddings corresponding to the events in the coarse subgraph, where \(\mathbf{f}_j\) represents the feature of the \(j\)-th video. We aim to identify the fine subset \(\mathcal{S'}\) of size \(n\) that maximizes the cumulative visual similarity with the current observation. The retrieved feature set \(\hat{\mathcal{F}}_n\) is formulated as:

\begin{equation}
\label{eq:fine_topn}
\begin{aligned}
\hat{\mathcal{F}}_n = \operatorname*{arg\,max}_{\substack{\mathcal{S'} \subset \mathcal{F}_{\text{sub}}, |\mathcal{S'}|=n}} \sum_{\mathbf{f} \in \mathcal{S'}} \operatorname{sim}_{\text{RAG}}(f_t^v, \mathbf{f}).
\end{aligned}
\end{equation}

Based on \(\hat{\mathcal{F}}_n\), we obtain the corresponding video set $\hat{\mathcal{V}}_\text{event}^{(n)}$ and event set $\hat{\mathcal{T}}_\text{event}^{(n)}$. This retrieved set provides a fine-grained knowledge prior \(\hat{\mathcal{G}}\) that combines visual and textual cues. The tail node event is then chosen by traversing the causal links among the retrieved events to guide high-level navigation decisions.





\subsection{Multimodal Knowledge Injection and Fusion}

Building on the fine-grained knowledge retrieval from the previous stage, where the top-\(n\) most relevant events and their corresponding video features are retrieved from \(\hat{\mathcal{G}}\), the fusion of this retrieved context into the agent’s current perception and instructions is achieved through two parallel pathways: Instruction Augmentation and Visual Enhancement.

\subsubsection{Instruction Augmentation}
To bridge the semantic gap in coarse-grained commands, we serialize the textual descriptions of retrieved events and append them to the original instruction $I$. Separated by a special token \texttt{<EVENT>}, this results in an enriched instruction $I'_t$ that explicitly provides the agent with missing procedural details:
\begin{equation}
I'_t = [I; \texttt{<EVENT>}; \hat{T}_{\text{event}}^{(1)}; \dots; \hat{T}_{\text{event}}^{(n)}].
\end{equation}

\subsubsection{Visual Enhancement via ASTFF}
Simultaneously, we introduce an ASTFF mechanism to enrich the agent's static perception with dynamic process knowledge. Instead of standard cross-attention, we design ASTFF as a Knowledge-Guided Transformer Block.
Specifically, we treat the agent's current panoramic observation $f_t^v$ as the Query ($Q$), seeking relevant visual cues from YE-KG. The retrieved video features $\hat{\mathcal{F}}_n$, which encode spatiotemporal dynamics of similar past events, serve as the Key ($K$) and Value ($V$). To explicitly distinguish between the static observation and dynamic video events in the shared latent space, learnable modality type tokens are prepended to the input features.
Through a multi-head attention mechanism followed by a residual connection and layer normalization, the module dynamically aligns the current view with the most relevant historical visual experiences. This yields an enhanced visual representation $f_t^{v'}$, formalized as:
\begin{equation}
\small
\begin{aligned}
  f^{v'}_t 
  &= \operatorname{ASTFF}(f_t^{v}, \hat{\mathcal{F}}_n) \\
  &= \operatorname{LayerNorm}\Bigl( f_t^v + \Bigl[ \operatorname*{Concat}_{h=1}^{H} \bigl( \text{Head}_h \bigr) \Bigr] W_O \Bigr),
\end{aligned}
\end{equation}
\noindent
where $\text{Head}_h = \operatorname{Attn}(f_t^v W_Q^h, \hat{\mathcal{F}}_n W_K^h, \hat{\mathcal{F}}_n W_V^h)$ represents the output of the $h$-th attention head.
Here, $W_Q, W_K, W_V$ and $W_O$ are learnable projections. This design enables the agent to adaptively incorporate the dynamic, temporal nature of the environment into its navigation strategy.

\subsection{Knowledge Injection in VLN}

We integrate our YE-KG into the vision-language navigation task to enrich coarse instructions and visual inputs with external knowledge. In a typical VLN scenario, an agent navigates an unseen indoor environment using a natural language instruction (e.g., “Find the blue sofa in the living room”). Such instructions can be semantically ambiguous or lacking in detail, making it challenging for the agent to identify the correct target. To address this, we augment the given instruction $I$ by appending a special token followed by relevant event knowledge retrieved from YE-KG, and we simultaneously enhance the agent’s visual observations with semantic cues from the knowledge graph. 
We formalize our knowledge-augmented VLN model as follows:
\begin{equation}
\left\{
\begin{aligned}
I'_t &= [I; \texttt{<EVENT>}; \hat{T}_{\text{event}}^{(1)}; \dots; \hat{T}_{\text{event}}^{(n)}]\\[4pt]
f_{t}^{v'} &= \operatorname{ASTFF}\!\bigl(f_t^{v},\,\{\hat{\mathcal{F}}_i\}_{i=1}^{n}\bigr), \\[4pt]
a_t        &= \pi\!\bigl(I_t',\,f_{t}^{v'}\bigr).
\end{aligned}
\right.
\end{equation}

At each time step $t$ the original instruction $I$ is extended to $I_t'$ by adding an \texttt{<EVENT>} or \texttt{<SCENE>} marker and a serialized list of knowledge entries $\{\hat{T}_\text{event}^{(i)}\}_{i=1}^{n}$ obtained by coarse-to-fine hierarchical retrieval. In parallel, the agent’s panoramic visual feature $\hat{\mathcal{F}}_n$ (extracted from its RGB-D observation) is fused with related visual cues ${v_k}$ via our ASTFF module, yielding an enriched visual representation $f_t^{v'}$. The VLN policy network $\pi$ then takes the augmented instruction $I_t'$ and the fused visual feature $f_{t}^{v'}$ as inputs to predict the next action $a_t$. 
\section{Experiment Result and Analysis}
\begin{table*}[!t]
\caption{\textbf{Performance comparison on the coarse-grained REVERIE~\cite{qi2020reverie} benchmark.} 
Our STE-VLN is implemented on top of the GOAT~\cite{wang2024vision} backbone. 
We report the validation and test metrics compared to state-of-the-art methods.}
\centering
\resizebox{\textwidth}{!}{
\begin{tabular}{@{} l l l l l l l l l l l l l @{}}
\toprule
\multirow{2}{*}{\textbf{Models}}
  & \multicolumn{4}{c}{\textbf{Validation Seen}}
  & \multicolumn{4}{c}{\textbf{Validation Unseen}}
  & \multicolumn{4}{c}{\textbf{Test Unseen}} \\
\cmidrule(lr){2-5} \cmidrule(lr){6-9} \cmidrule(lr){10-13}
  & \textbf{SR$\uparrow$} & \textbf{SPL$\uparrow$} & \textbf{RGS$\uparrow$} & \textbf{RGSPL$\uparrow$}
  & \textbf{SR$\uparrow$} & \textbf{SPL$\uparrow$} & \textbf{RGS$\uparrow$} & \textbf{RGSPL$\uparrow$}
  & \textbf{SR$\uparrow$} & \textbf{SPL$\uparrow$} & \textbf{RGS$\uparrow$} & \textbf{RGSPL$\uparrow$} \\
\midrule
HAMT~\cite{chen2021history} & 43.29 & 40.19 & 27.2 & 25.18 & 32.95 & 30.2 & 18.92 & 17.28 & 30.40 & 26.67 & 14.88 & 13.08 \\
HOP+~\cite{qiao2023hop+} & 55.87 & 49.55 & 40.76 & 36.22 & 36.07 & 31.13 & 22.49 & 19.33 & 33.82 & 28.24 & 20.20 & 16.86 \\
DSRG~\cite{wang2023dual} & 75.69 & 68.09 & 61.07 & 54.72 & 47.83 & 34.02 & 32.69 & 23.37 & 54.04 & 37.09 & 32.49 & 22.18 \\
GridMM~\cite{wang2023gridmm} & -- & -- & -- & -- & 51.37 & 36.47 & 34.57 & 24.56 & 53.13 & 36.60 & 34.87 & 23.45 \\
DUET~\cite{chen2022think} & 71.75 & 63.94 & 57.41 & 51.14 & 46.98 & 33.73 & 32.15 & 23.03 & 52.51 & 36.06 & 31.88 & 22.06 \\
BEVBert~\cite{an2022bevbert} & 73.72 & 65.32 & 57.7 & 51.73 & 51.78 & 36.37 & 34.71 & 24.44 & 52.81 & 36.41 & 32.06 & 22.09 \\
GOAT~\cite{wang2024vision} & 78.64 & 71.4 & 63.74 & 57.85 & 53.37 & 36.7 & 38.43 & 26.09 & 57.72 & 40.53 & 38.32 & 26.70 \\
\rowcolor{lightgray}
\textbf{STE-VLN (Ours)} & \textbf{80.65}\textsubscript{↑2.01}
                    & \textbf{71.20}\textsubscript{↓0.20}
                    & \textbf{64.52}\textsubscript{↑0.78}
                    & \textbf{57.92}\textsubscript{↑0.07}
                    & \textbf{55.33}\textsubscript{↑1.96}
                    & \textbf{36.46}\textsubscript{↓0.24}
                    & \textbf{39.92}\textsubscript{↑1.49}
                    & \textbf{26.12}\textsubscript{↑0.03}
                    & \textbf{59.55}\textsubscript{↑1.83}
                    & \textbf{40.19}\textsubscript{↓0.34}
                    & \textbf{39.75}\textsubscript{↑1.43}
                    & \textbf{26.62}\textsubscript{↓0.08} \\
\bottomrule
\end{tabular}
}
\label{tab:reverie_results}
\end{table*}

\begin{table}[!t]
\centering
\caption{\textbf{Performance comparison on the fine-grained R2R~\cite{anderson2018vision} benchmark.} 
Our STE-VLN is implemented on top of the GOAT~\cite{wang2024vision} backbone, demonstrating the method's generalization capability on fine-grained instructions.}
\renewcommand{\arraystretch}{0.95}
\resizebox{0.95\columnwidth}{!}{%
\begin{tabular}{l l l l l l}
\toprule
\multirow{2}{*}{\textbf{Models}} &
\multicolumn{2}{c}{\textbf{Validation Seen}} &
\multicolumn{2}{c}{\textbf{Validation Unseen}} \\
\cmidrule(lr){2-3}\cmidrule(lr){4-5}
& \textbf{SR$\uparrow$} & \textbf{OSR$\uparrow$}
& \textbf{SR$\uparrow$} & \textbf{OSR$\uparrow$} \\
\midrule
HAMT~\cite{chen2021history} & 76 & 82 & 66 & 73 \\
DUET~\cite{chen2022think} & 79 & 86 & 72 & 81 \\
TD-STP~\cite{zhao2022target} & 77 & 83 & 70 & 76 \\
GeoVLN~\cite{huo2023geovln} & 79 & -- & 68 & -- \\
DSRG~\cite{wang2023dual} & 81 & 88 & 73 & 81 \\
GridMM~\cite{wang2023gridmm} & -- & -- & 75 & -- \\
GELA~\cite{Cui_2023_ICCV} & 76 & -- & 71 & -- \\
EnvEdit~\cite{li2022envedit1} & 77 & -- & 69 & -- \\
BEVBert~\cite{an2022bevbert} & 81 & 88 & 75 & 84 \\
GOAT~\cite{wang2024vision} & 83.74 & 88.64 & 77.82 & 84.72 \\
\rowcolor{lightgray}
STE-VLN (Ours) & \textbf{84.95}$_{\uparrow 1.21}$ & \textbf{89.84}$_{\uparrow 1.20}$ & \textbf{79.01}$_{\uparrow 1.19}$ & \textbf{85.90}$_{\uparrow 1.18}$ \\
\bottomrule
\end{tabular}%
}
\label{tab:r2r_results}
\end{table}

\begin{table}[!t]
\centering
\renewcommand{\arraystretch}{0.95} 
\caption{\textbf{Performance comparison on the continuous environment R2R-CE~\cite{krantz2020beyond} benchmark.} 
Our STE-VLN is implemented on top of the ETPNav~\cite{an2024etpnav} backbone, validating its effectiveness in continuous spaces.}
\resizebox{\columnwidth}{!}{%
\begin{tabular}{l c c c c c c c c}
\toprule
\multirow{2}{*}{\textbf{Methods}} &
\multicolumn{4}{c}{\textbf{Validation Seen}} &
\multicolumn{4}{c}{\textbf{Validation Unseen}} \\
\cmidrule(lr){2-5}\cmidrule(lr){6-9}
& \textbf{SR$\uparrow$} & \textbf{SPL$\uparrow$} & \textbf{NE$\downarrow$} & \textbf{OSR$\uparrow$}
& \textbf{SR$\uparrow$} & \textbf{SPL$\uparrow$} & \textbf{NE$\downarrow$} & \textbf{OSR$\uparrow$} \\
\midrule
Seq2Seq~\cite{krantz2020beyond}     & 37 & 35 & 7.12 & 46 & 32 & 30 & 7.37 & 40 \\
SASRA~\cite{irshad2021sasra}      & 36 & 34 & 7.71 & -- & 24 & 22 & 8.32 & -- \\
CMTP~\cite{chen2021topological}       & 36 & 31 & 7.10 & 56 & 26 & 23 & 7.90 & 38 \\
LAW~\cite{raychaudhuri2021language}        & 40 & 37 & --   & -- & 35 & 31 & --   & -- \\
HPN~\cite{krantz2021waypoint}          & 46 & 43 & 5.48 & 53 & 36 & 34 & 6.31 & 40 \\
CM2~\cite{georgakis2022cross}          & 43 & 35 & 6.10 & 51 & 34 & 28 & 7.02 & 42 \\
WS-MGMAP~\cite{chen2022weakly}     & 47 & 43 & 5.65 & 52 & 39 & 34 & 6.28 & 48 \\
CWP-CMA~\cite{hong2022bridging}      & 51 & 45 & 5.20 & 61 & 41 & 36 & 6.20 & 52 \\
CWP-RecBERT~\cite{hong2022bridging}  & 50 & 44 & 5.02 & 59 & 44 & 39 & 5.74 & 53 \\
Sim2Sim~\cite{krantz2022sim}      & 52 & 44 & 4.67 & 61 & 43 & 36 & 6.07 & 52 \\
ETPNav~\cite{an2024etpnav}      & 66 & 59 & 3.95 & 72 & 59 & 49 & 4.71 & 65 \\
\rowcolor{gray!25}
STE-VLN (Ours) & \textbf{68} & \textbf{60} & \textbf{3.82} & \textbf{74} & \textbf{61} & \textbf{50} & \textbf{4.57} & \textbf{66} \\
\bottomrule
\end{tabular}%
}
\label{tab:r2r_ce_results}
\end{table}

\begin{table*}[!t]
\centering
\renewcommand{\arraystretch}{0.95} 
\caption{
\textbf{Ablation study on different combinations of event and scene knowledge.}
``TOP-$n$'' is the number of fine-grained knowledge items. ``EVENT'' counts descriptions labeled \texttt{Event-0}, while ``SCENE'' counts those labeled \texttt{Scene-1}.}
\resizebox{0.65 \textwidth}{!}{%
\begin{tabular}{c c c c c c c c c c c}
\toprule
\multirow{2}{*}{\textbf{TOP-$n$}} &
\multirow{2}{*}{\textbf{EVENT}} &
\multirow{2}{*}{\textbf{SCENE}} &
\multicolumn{4}{c}{\textbf{Validation Seen}} &
\multicolumn{4}{c}{\textbf{Validation Unseen}} \\
\cmidrule(lr){4-7} \cmidrule(lr){8-11}
    & & & \textbf{SR} & \textbf{SPL}
    & \textbf{RGS} & \textbf{RGSPL}
    & \textbf{SR} & \textbf{SPL}
    & \textbf{RGS} & \textbf{RGSPL} \\
\midrule
0 & 0 & 0 & 78.64 & 71.40 & 63.74 & 57.85 & 53.37 & 36.70 & 38.43 & 26.09 \\
\midrule
\multirow{2}{*}{1} 
    & 1 & 0 & 80.15 & 71.46 & 64.07 & 57.81 & 54.55 & 36.77 & 39.08 & 26.07 \\
    & 0 & 1 & 78.14 & 71.24 & 63.48 & 57.90 & 52.87 & 36.50 & 38.04 & 26.11 \\
\midrule
\multirow{3}{*}{2} 
    & 2 & 0 & 80.60 & 71.43 & \textbf{64.55} & 57.82 & 55.10 & 36.75 & \textbf{40.01} & 26.07 \\
    & 1 & 1 & 80.55 & 71.27 & 64.46 & 57.87 & 54.98 & 36.53 & 39.69 & 26.10 \\
    & 0 & 2 & 77.71 & 71.09 & 63.39 & 57.94 & 52.64 & 36.40 & 37.90 & 26.13 \\
\midrule
    & 3 & 0 & 80.14 & 71.54 & 64.21 & 57.78 & 55.01 & 36.86 & 39.38 & 26.06 \\
\rowcolor{gray!25} 
\cellcolor{white}
    & 2 & 1 & 
    \textbf{80.65} & 71.20 &
    64.52 & 57.92 &
    \textbf{55.33} & 36.46 &
    39.92 & 26.12 \\
    & 1 & 2 & 79.95 & 71.20 & 64.09 & \textbf{57.93} & 54.83 & 36.45 & 39.08 & 26.12 \\
\multirow{-4}{*}{3} 
    & 0 & 3 & 77.49 & 71.11 & 63.22 & 57.92 & 52.51 & 36.40 & 37.48 & 26.12 \\
\bottomrule
\end{tabular}%
}
\label{tab:Event_Knowledge_vs.Scene–Entity_Knowledge}
\end{table*}

\subsection{Experimental Setup}

\subsubsection{Datasets}
We evaluate our approach on two representative Vision-Language Navigation benchmarks built upon the Matterport3D simulator~\cite{chang2017matterport3d}. REVERIE~\cite{qi2020reverie} offers a \emph{coarse--grained} reference expression benchmark comprising 10{,}318 panoramas collected from 86 buildings, 4{,}140 target objects across 489 categories, and 21{,}702 crowd‑sourced instructions with an average length of 18~tokens. Since instructions in REVERIE typically refer to remote objects without detailed path descriptions, this dataset heavily tests the agent's ability to infer high-level semantic plans.
In contrast, R2R~\cite{anderson2018vision} is a fine-grained navigation corpus containing 7{,}189 templated trajectories spanning 90 real‑world indoor scenes, paired with roughly 22\,k human written instructions whose mean length is 29~tokens. This dataset requires the agent to strictly follow step-by-step guidance.
Additionally, to assess robustness in continuous control settings, we utilize R2R-CE~\cite{krantz2020beyond}, which lifts the graph-based constraints and requires the agent to predict low-level velocities in a continuous 3D space.

\subsubsection{Evaluation Metrics}

As in~\cite{an2024etpnav,wang2024vision,an2022bevbert}, we adapt the following navigation metrics.
For REVERIE we report Success Rate (SR), Success weighted by Path Length (SPL), Remote Grounding Success (RGS) and its path‑length‑aware variant (RGSPL). On R2R and R2R-CE, we follow the standard protocol reporting SR, SPL, Navigation Error (NE), and Oracle Success Rate (OSR).

\subsubsection{Implementation Details}
We implement our STE-VLN framework on top of state-of-the-art backbones: GOAT~\cite{wang2024vision} for the REVERIE~\cite{qi2020reverie} and R2R~\cite{anderson2018vision} benchmarks, and ETPNav~\cite{an2024etpnav} for the R2R-CE~\cite{krantz2020beyond} benchmark. 
Following standard protocols~\cite{wang2024vision,an2024etpnav}, pre-training employs Masked Language Modeling (MLM), Single-Step Action Prediction (SAP), and Context-Action Fidelity Prediction (CFP) objectives for R2R. For REVERIE, we additionally include the Object Grounding (OG) objective. We adapt EnvEdit data augmentation~\cite{li2022envedit} utilizing two synthetic corpora~\cite{hao2020towards,wang2023res}.
Pre-training is conducted on four NVIDIA 4090 GPUs using the AdamW optimizer ($\eta{=}5\times10^{-5}$, batch size 48) for 300\,k iterations. During fine-tuning, sentence embeddings are extracted via \textsc{Sentence-Transformer}~\cite{reimers2019sentence} for retrieval, while frame-level visual tokens are encoded by a Vision Transformer (ViT)~\cite{dosovitskiy2020image}. For the ASTFF module, we implement it as a single-layer Transformer block with the number of attention heads $H=8$ and a hidden dimension of $d=768$. We set the batch size to 10, the learning rate to $2\times10^{-5}$, and train for up to 50\,k iterations.

\subsection{Comparisons with State-of-the-Arts}


To validate the universality and effectiveness of our proposed approach, we evaluate the performance across three representative benchmarks: REVERIE, R2R, and R2R-CE. 
Our experimental strategy involves integrating the STE-VLN framework into current state-of-the-art (SOTA) backbones to demonstrate that our explicit event knowledge can push the performance boundaries of existing top-performing models.
As reported in Tables~\ref{tab:reverie_results},~\ref{tab:r2r_results}, and~\ref{tab:r2r_ce_results}, our method establishes new state-of-the-art records across three datasets.
On the challenging REVERIE dataset, by equipping GOAT~\cite{wang2024vision} with our YE-KG, we achieve an impressive average SR improvement of 1.93\%.
Specifically, on the test-unseen split, our method reaches an SR of 59.55\% (+1.83\%) and 55.33\% on the val-unseen split (+1.96\%). 
These consistent gains confirm that our method functions as a generalized performance booster for diverse navigation architectures.

\subsubsection{\textbf{Coarse‑grained Instruction Comparisons (REVERIE)}}



Table~\ref{tab:reverie_results} presents the comparative results on the REVERIE benchmark. Distinguished by its coarse-grained instructions, this dataset poses a significant challenge as it requires agents to perform high-level reasoning to locate remote objects without step-by-step trajectory guidance. This setting serves as an ideal testbed to validate our method's capability in resolving instruction ambiguity.

By integrating our STE-VLN framework into the strong SOTA baseline GOAT~\cite{wang2024vision}, we achieve a new state-of-the-art record. Specifically, on the validation unseen split, our method boosts the Success Rate (SR) to 55.33\%, surpassing the vanilla GOAT (53.37\%) by an absolute margin of 1.96\%. 
Crucially, this improvement stems from the explicit indoor navigation priors provided by our YE-KG. When facing ambiguous commands (e.g., ``find the sink''), the retrieved event priors guide the agent to explore semantically relevant areas (e.g., ``navigating to a bathroom'') rather than wandering aimlessly. This is evidenced by the consistent gain in Remote Grounding Success (RGS), which increases to 39.92\% (+1.49\%), verifying that our agent better understands the room-to-object associations.

It is worth noting that while SR improves significantly, SPL shows a relatively modest increase. This trend is consistent with the nature of knowledge-driven navigation: the retrieved priors prioritize reaching the correct goal by encouraging necessary exploration of relevant context, ensuring success even if it occasionally incurs a slightly longer trajectory.

\begin{figure*}[!t]
    \centering
    \includegraphics[width=0.85\textwidth]{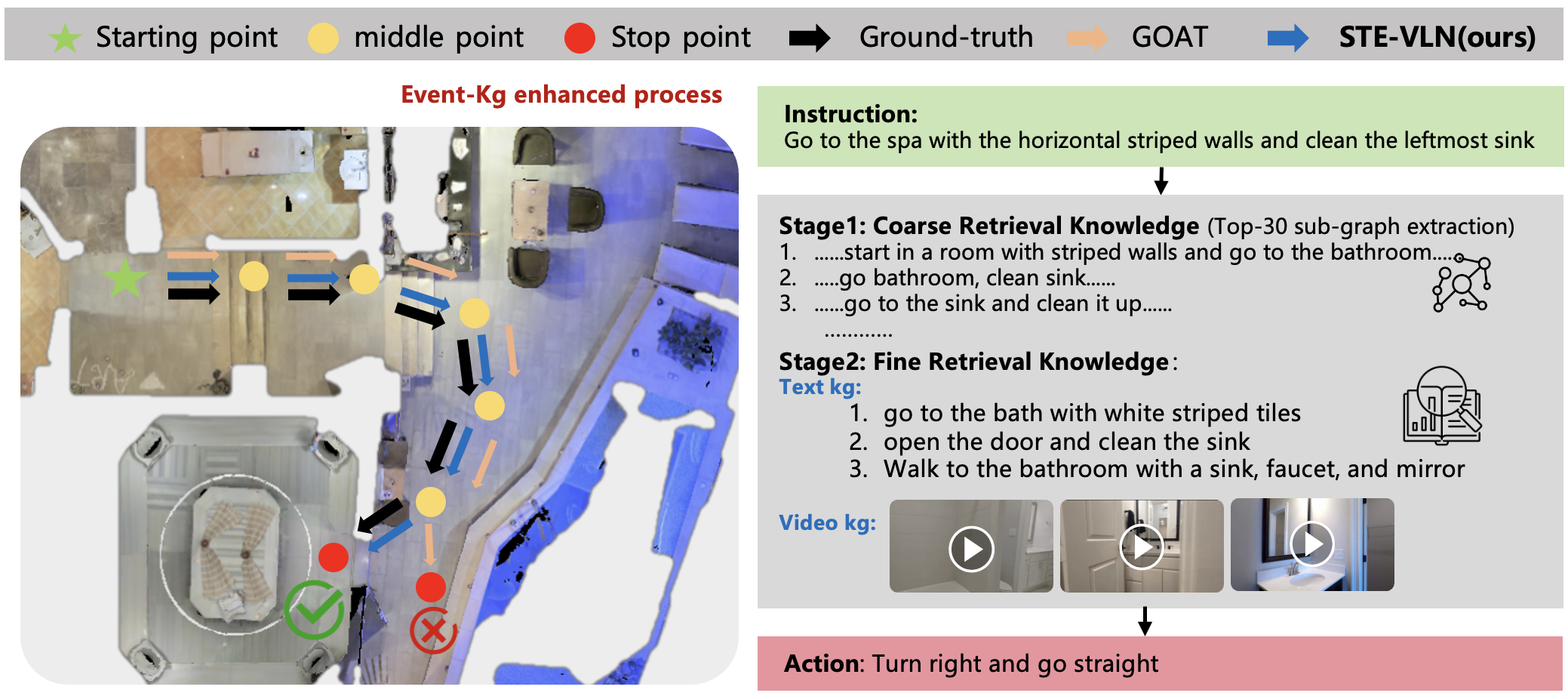}
    \caption{Visualization of event-level knowledge–guided decision correction in unseen VLN scenarios.
    By retrieving related event sequences and visual cues via a coarse-to-fine mechanism, our method compensates for missing procedural priors and guides the agent to the target. The light orange arrows represent the trajectories of the GOAT~\cite{wang2024vision} baseline, which erroneously leads to a bathtub instead of the target sink.}
    \label{fig:case_study}
\end{figure*}

Fig.~\ref{fig:case_study} qualitatively compares GOAT with our STE-VLN model on an unseen scene. The instruction requires the agent to ``go to the spa... and clean the leftmost sink.'' The baseline GOAT agent fails because it lacks the prior knowledge of what a ``spa with a sink'' typically looks like or how to approach it. In contrast, our model performs a coarse retrieval to extract a sub-graph of related events (e.g., ``go to bathroom, clean sink''). During navigation, the Fine Retrieval mechanism retrieves specific video clips of ``opening a door'' and ``cleaning a sink.'' These explicit visual priors guide the agent to correctly identify the bathroom entrance and navigate to the target sink.
\subsubsection{\textbf{Fine‑grained Instruction Comparisons (R2R)}}

Table \ref{tab:r2r_results} further demonstrates the generalization of STE-VLN in the fine-grained R2R task. We utilize the current state-of-the-art model GOAT~\cite{wang2024vision} as our backbone. In the table, the entry STE-VLN (Ours) refers to the GOAT model augmented with our proposed event knowledge retrieval and fusion mechanism. Despite the availability of detailed instructions in R2R, incorporating our framework continues to deliver clear gains. On the challenging R2R val-unseen split, our method lifts the SR of the GOAT backbone from 77.82\% to 79.01\% (+1.19\%), and the Oracle Success Rate (OSR) from 84.72\% to 85.90\% (+1.18\%). These improvements suggest that even when detailed instructions are available, the agent benefits from the visual foresight provided by the retrieved video events, helping it to resolve local ambiguities better than the original SOTA model.

\subsubsection{\textbf{Continuous Environment Comparisons (R2R-CE)}}


To further verify robustness in realistic, robotic-like settings, we conduct experiments on the R2R-CE benchmark. We compare our framework directly with the state-of-the-art continuous planner ETPNav~\cite{an2024etpnav}. In Table \ref{tab:r2r_ce_results}, STE-VLN (Ours) denotes the implementation where our framework is integrated into the ETPNav backbone.
The results show that integrating STE-VLN significantly enhances the robustness of the continuous planner. Compared to the vanilla ETPNav, our method improves the SR on the Validation Unseen split from 59\% to 61\% and SPL from 49\% to 50\%. On Validation Seen, we explicitly boost the SR from 66\% to 68\%. This indicates that high-level event planning remains effective even when the agent must handle complex low-level motion control, providing a stable global goal that helps the baseline agent recover from local collision-avoidance maneuvers.

\begin{figure*}
    \centering
    \includegraphics[width=0.9\linewidth]{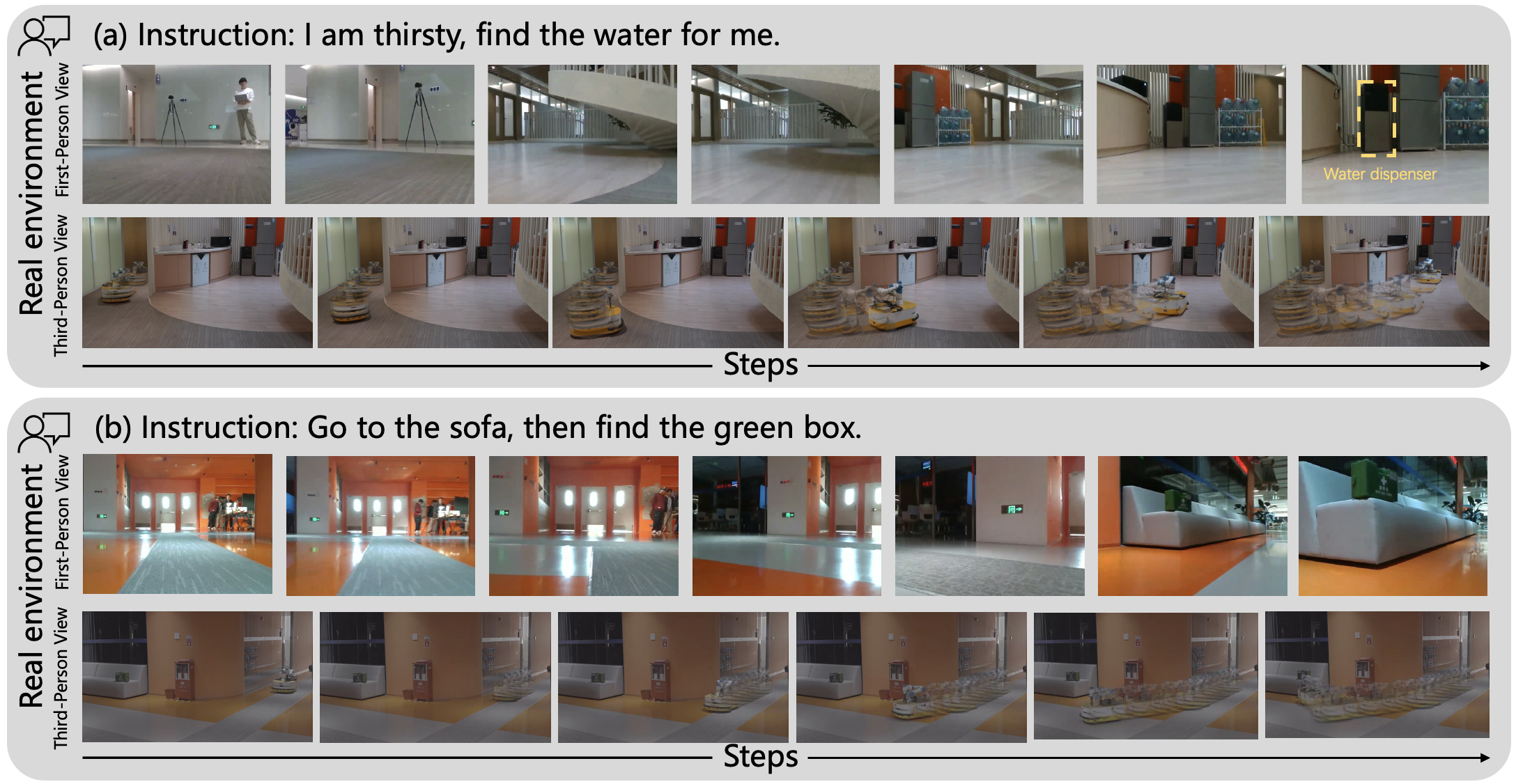}
    \caption{Real-world deployment. The agent robustly follows natural language instructions in real office settings. (a) ``I am thirsty, find the water for me'': navigation to the water dispenser. (b) ``Go to the sofa, then find the green box'': localization of the sofa followed by retrieval of the green box. Top rows: first-person views; bottom rows: third-person views, illustrating seamless traversal of diverse functional areas.}
    \label{fig:real_world}
\end{figure*}
\begin{table}[t]
\centering
\renewcommand{\arraystretch}{0.954} 
\caption{\textbf{Ablation study results on different fusion strategies.}}
\resizebox{\columnwidth}{!}{%
\begin{tabular}{c c c c c c c c c c}
\toprule
\multicolumn{2}{c}{\textbf{Fusion Strategy}} &
\multicolumn{4}{c}{\textbf{Validation Seen}} &
\multicolumn{4}{c}{\textbf{Validation Unseen}} \\
\cmidrule(lr){1-2}\cmidrule(lr){3-6}\cmidrule(lr){7-10}
\textbf{Text} & \textbf{Visual} &
\textbf{SR$\uparrow$} & \textbf{SPL$\uparrow$} &
\textbf{RGS$\uparrow$} & \textbf{RGSP$\uparrow$} &
\textbf{SR$\uparrow$} & \textbf{SPL$\uparrow$} &
\textbf{RGS$\uparrow$} & \textbf{RGSP$\uparrow$} \\
\midrule
\xmark & \xmark & 78.64 & 71.40 & 63.74 & 57.85 & 53.37 & 36.70 & 38.43 & 26.09 \\
\cmark & \xmark & 80.32 & 71.28 & 64.47 & 57.91 & 54.90 & 36.54 & 39.80 & 26.12 \\
\xmark & \cmark & 79.34 & 71.32 & 63.98 & 57.81 & 53.95 & 36.59 & 38.82 & 26.07 \\
\rowcolor{gray!25}
\cmark & \cmark &
\textbf{80.65} & \textbf{71.20} & \textbf{64.52} & \textbf{57.92} &
\textbf{55.33} & \textbf{36.46} & \textbf{39.92} & \textbf{26.12} \\
\bottomrule
\end{tabular}%
}
\label{tab:Textual_KG_vs.Visual_KG}
\end{table}

\begin{table}[t]
\caption{\textbf{Efficiency Analysis.} The computational cost introduced by STE-VLN is negligible compared to the backbone model. Latency is measured on a single NVIDIA 4090 GPU.}
\centering
\small
\renewcommand{\arraystretch}{1.15}
\setlength{\tabcolsep}{6pt}
\begin{tabular}{l l c}
\toprule
\textbf{Component} & \textbf{Metric} & \textbf{Value} \\
\midrule
\multirow{2}{*}{Model Overhead} & ASTFF Params & 4.73 M \\
                                & KG Features Storage & 487 MB \\
\midrule
\multirow{2}{*}{Inference Latency} & Coarse Retrieval (Once) & 3.92 ms \\
                                   & Fine Retrieval (Per Step) & 0.02 ms \\
\bottomrule
\end{tabular}
\label{tab:efficiency}
\end{table}

\subsection{Ablation study}

We investigate the impact of different knowledge components in Table~\ref{tab:Event_Knowledge_vs.Scene–Entity_Knowledge} and Table~\ref{tab:Textual_KG_vs.Visual_KG}.

\textbf{Event vs. Scene Knowledge:} Table~\ref{tab:Event_Knowledge_vs.Scene–Entity_Knowledge} reveals a compelling trade-off. First, the baseline performance without any external knowledge (0 Event + 0 Scene) represents the standard reactive navigation paradigm, which yields the lowest Success Rate (SR) of 53.37\% on val-unseen. Increasing the number of event nodes (which describe dynamic transitions like ``walk into...'') consistently improves the Success Rate (SR), confirming that process knowledge is crucial for planning. However, purely using events might lead to slight overshooting. Adding a single \emph{scene} node (which describes static details like ``a room with a red curtain'') provides a grounding signal that helps the agent verify its final destination. Consequently, the optimal configuration is 2 Events + 1 Scene, which yields the highest val-unseen SR of 55.33\%.

\textbf{Multimodal Fusion:} The ablation in Table~\ref{tab:Textual_KG_vs.Visual_KG} demonstrates the necessity of multimodal injection. Removing the textual branch (which provides semantic event cues to the instruction encoder) drops the SR to 53.95\%. Removing the visual branch (which provides spatiotemporal cues via ASTFF) drops it to 54.90\%. Re-activating both pathways achieves the best performance of \textbf{55.33\%}. This confirms that text provides the plan, while video features provide the visual intuition, and both are essential for robust navigation.

\subsection{Real-World Deployment}

To demonstrate the practical utility of our framework beyond simulation, we deploy the STE-VLN model on a physical mobile robot named ``NXROBO Leo'' (Fig.~\ref{fig:real_world}). The robot is equipped with a LiDAR for obstacle avoidance and an RGB-D camera for visual perception. We map a real-world office environment and test the agent with natural language commands such as ``Find me a water dispenser''
Despite the domain gap between the Matterport3D simulator and the real office, our agent successfully plan a path from the corridor to the pantry. As shown in Fig.~\ref{fig:real_world}, the robot effectively recognize the transition from the hallway to the functional area with water dispensers. This successful Sim-to-Real transfer can be attributed to the \textbf{generalizable navigation paradigms} learned from the YE-KG. The diverse, open-world video data in our graph provides the agent with robust visual concepts (e.g., what a ``pantry entrance'' looks like in various lighting conditions) that generalize better than training solely on limited simulator data.

\subsection{Efficiency Analysis}

We analyze the computational overhead of our proposed framework in Table~\ref{tab:efficiency}. The introduced ASTFF fusion module is lightweight, adding only 4.73M parameters, which is a fraction of the total model size. Regarding storage, the pre-extracted multimodal features of YE-KG occupy 487 MB. This compact size allows the entire knowledge base to be loaded into the VRAM of modern GPUs, enabling ultra-fast access. 
In terms of latency, the retrieval process is highly efficient: the Coarse-grained Retrieval (executed once per instruction) takes approximately 3.92 ms, and the Fine-grained Retrieval (executed at each navigation step) takes only 0.02 ms. These results confirm that our method improves performance with negligible computational cost, making it highly suitable for real-time deployment on robotic platforms with limited resources.

\section{Conclusion}

In this paper, we propose a systematic framework to address the cognitive deficit of explicit process knowledge in VLN agents. To the best of our knowledge, we construct YE-KG, the first large-scale multimodal knowledge graph derived from real-world indoor videos, capable of encoding generalizable navigation patterns. For navigation, we introduce STE-VLN, which leverages a Coarse-to-Fine Hierarchical Retrieval mechanism to dynamically fuse textual and visual event priors into the policy.
Experiments on REVERIE, R2R, and R2R-CE demonstrate that our event-centric paradigm yields consistent improvements, outperforming state-of-the-art approaches across diverse action spaces. Furthermore, our successful real-world deployment verifies the practical efficacy of our approach in bridging the Sim-to-Real gap. In the future, we plan to further explore grounding embodied planning in explicit, multimodal event knowledge, paving the way for agents that truly understand the causal dynamics of the physical world.



\bibliographystyle{IEEEtran}  
\bibliography{refs}  

\end{document}